\documentclass{SCIS2021}
\usepackage[table]{xcolor}
\usepackage{multirow}
\usepackage{pgfplots}
\usepackage{CJKutf8}
\newcommand{\tabincell}[2]{\begin{tabular}{@{}#1@{}}#2\end{tabular}}
\newcommand{\cmark}{\ding{51}}
\newcommand{\xmark}{\ding{55}}

\begin{document}
\ArticleType{RESEARCH PAPER}
\Year{2020}
\Month{}
\Vol{}
\No{}
\DOI{}
\ArtNo{}
\ReceiveDate{}
\ReviseDate{}
\AcceptDate{}
\OnlineDate{}

\title{CPT: A Pre-Trained Unbalanced Transformer for Both Chinese Language Understanding and Generation}{CPT: A Pre-Trained Unbalanced Transformer for Both Chinese Language Understanding and Generation}

\author[1,3]{Yunfan Shao}{}
\author[1,3]{Zhichao Geng}{}
\author[1,3]{Yitao Liu}{}
\author[1,3]{Junqi Dai}{}
\author[1,3]{Hang Yan}{}
\author[2]{\\Fei Yang}{}
\author[2]{Li Zhe}{}
\author[2]{Hujun Bao}{}
\author[1,3]{Xipeng Qiu}{{xpqiu@fudan.edu.cn}}

\AuthorMark{Shao Y F}

\AuthorCitation{Shao Y F, Geng Z C, Liu Y T, et al}


\address[1]{School of Computer Science, Fudan University, Shanghai {\rm 200433}, China}
\address[2]{Zhejiang Lab, Hangzhou {\rm 311121}, China}
\address[3]{Shanghai Key Laboratory of Intelligent Information Processing, Fudan University, Shanghai {\rm 200433}, China}

\abstract{In this paper, we take the advantage of previous pre-trained models (PTMs) and propose a novel \textbf{C}hinese \textbf{P}re-trained Unbalanced \textbf{T}ransformer (\textbf{CPT}).
Different from previous Chinese PTMs, CPT is designed to utilize the shared knowledge between natural language understanding (NLU) and natural language generation (NLG) to boost the performance.
CPT consists of three parts: a shared encoder, an understanding decoder, and a generation decoder. Two specific decoders with a shared encoder are pre-trained with masked language modeling (MLM) and denoising auto-encoding (DAE) tasks, respectively.
With the partially shared architecture and multi-task pre-training, CPT can (1) learn specific knowledge of both NLU or NLG tasks with two decoders and (2) be fine-tuned flexibly that fully exploits the potential of the model.
Moreover, the unbalanced Transformer saves the computational and storage cost, which makes CPT competitive and greatly accelerates the inference of text generation.
Experimental results on a wide range of Chinese NLU and NLG tasks show the effectiveness of CPT\footnote{Code is available at https://github.com/fastnlp/CPT}.}

\keywords{pre-trained model, transformer, language model, generation, unified model}

\maketitle

\section{Introduction}
Recently, large-scale pre-trained models (PTMs) have become backbone models for many natural language processing (NLP) tasks~\cite{qiu20ptm}. However, existing PTMs are usually trained with different architectures and pre-training tasks. When applying PTMs to a downstream task, we should choose a suitable one as the backbone model according to its pre-training nature. For example, we usually select BERT or RoBERTa ~\cite{Devlin19bert,Liu19roberta} as the backbone model for natural language understanding (NLU) tasks, and BART or GPT ~\cite{Lewis20BART,Radford18gpt} for natural language generation (NLG) tasks. With the success of PTMs in English, many works have been done to train the counterparts for Chinese ~\cite{Cui19cnbert,Sun19ernie,Wei19nezha,Zhang20cpm,Zhang21cpm2,Huawei21Pangu}. However, these Chinese PTMs usually follow the settings of English PTMs, which makes these models focus on either language understanding or language generation, lacking the use of sharing knowledge between NLU and NLG tasks.
Therefore, it is attractive to pre-train a joint model for both NLU and NLG tasks.

Few works attempt to fuse NLU and NLG into a unified model. UniLMs~\cite{Dong19UniLM,Bao20UniLMv2} and GLM~\cite{Du21GLM}  adapt a unified Transformer encoder for both understanding and generation; however, their architectures restrict them to employ more flexible pre-training tasks, such as denoising auto-encoding (DAE) used in BART, a widely successful pre-training task for NLG. PALM~\cite{Bi20PALM} adopts the standard Transformer and adds an auxiliary masked language modeling (MLM) task to enhance the understanding ability; however, it still focuses on language generation tasks.

\begin{figure*}[t!]
    \centering
  \begin{minipage}[t]{0.3\textwidth}
        \centering
        \includegraphics[page=2,width=0.5\textwidth]{cpt-pretrain-crop.pdf}
  \end{minipage}
  \begin{minipage}[t]{0.53\textwidth}
    \centering
   
    \includegraphics[page=4,width=0.78\textwidth]{cpt-pretrain-crop.pdf}
  \end{minipage}
  \\
  \begin{minipage}[t]{0.3\textwidth}
    \centering
    \caption*{(a) BERT}
    \label{fig:bert}
  \end{minipage}
  \begin{minipage}[t]{0.53\textwidth}
    \centering
    \caption*{(b) BART}
    \label{fig:bart}
  \end{minipage}
  \begin{minipage}[t]{0.3\textwidth}
        \centering
        \includegraphics[page=3,width=0.5\textwidth]{cpt-pretrain-crop.pdf}
  \end{minipage}
  \begin{minipage}[t]{0.53\textwidth}
    \centering
    \includegraphics[page=1,width=\textwidth]{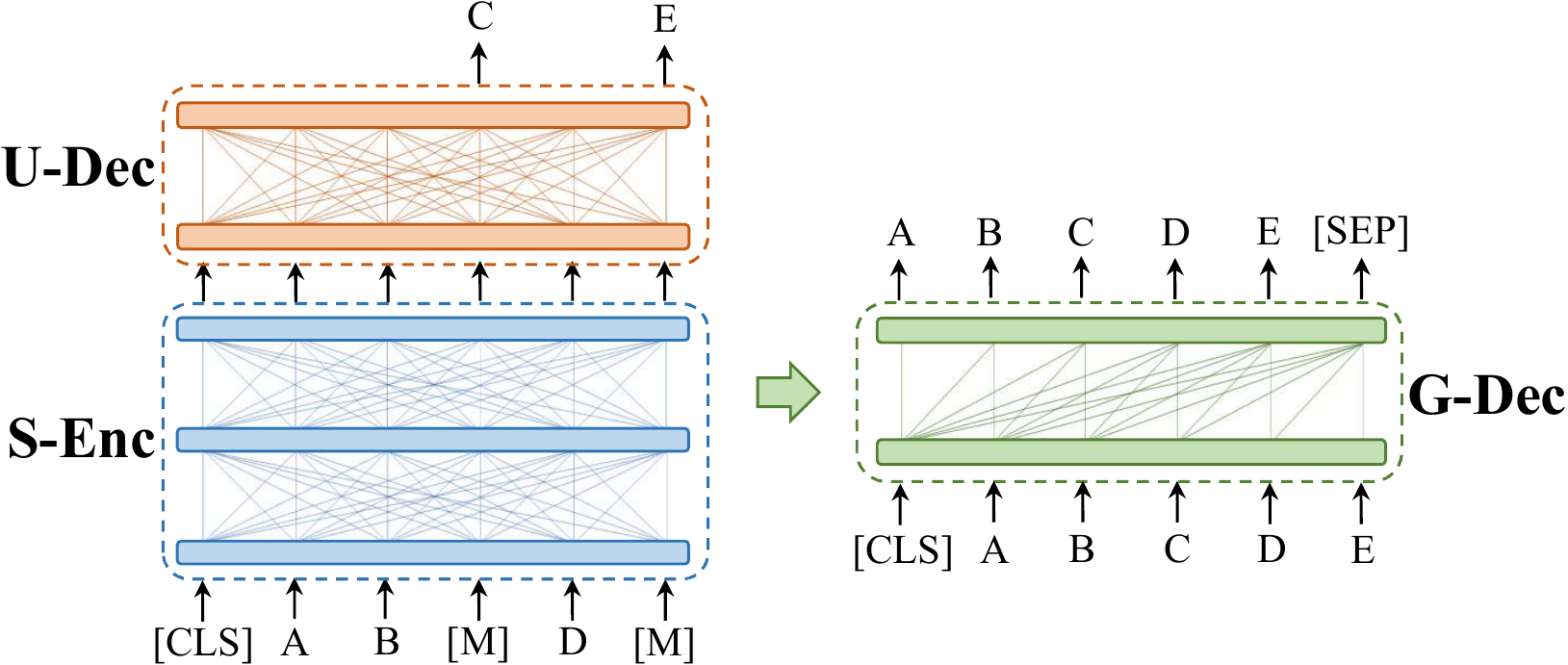}
  \end{minipage}
  \\
  \begin{minipage}[t]{0.3\textwidth}
    \centering
    \caption*{(c) GPT}
    \label{fig:gpt}
    \end{minipage}
  \begin{minipage}[t]{0.53\textwidth}
    \centering
    \caption*{(d) CPT}
    \label{fig:cpt}
  \end{minipage}
  \caption{Architecture of CPT and the counterpart PTMs. Different from other PTMs, CPT consists of three parts: a shared encoder (\textbf{S-Enc}), an understanding decoder (\textbf{U-Dec}) and a generation decoder (\textbf{{G-Dec}}).}
  \label{fig:pretrain}
\end{figure*}

In this paper, we propose \textbf{CPT}, a novel \textbf{C}hinese \textbf{P}re-trained Unbalanced \textbf{T}ransformer for both NLU and NLG tasks. The architecture of CPT is very concise (as shown in Figure~\ref{fig:pretrain}), which divides a full Transformer encoder-decoder into three parts:  1) a shared encoder to capture the common representation; 2) a decoder for understanding, which uses full self-attention and is pre-trained with masked language modeling (MLM); 3) a decoder for generation, which adopts masked self-attention and is pre-trained with the DAE task. By multi-task pre-training, CPT is able to improve the performance on both language understanding and generation, respectively.

\begin{table*}[ht!]
  \centering
    \tiny
    \caption{Summary of some representative Chinese PTMs. ``\# Params" refers to the number of parameters. ``Arch." refers to the model architecture. ``LM'' refers to language modeling in auto-regression fashion, while ``Seq2Seq MLM'' refers to masked language modeling in Seq2Seq fashion.
     ``Tok.", ``Masking" and ``Prediction" refer to the tokenization, masking and prediction granularity of the model, respectively.  ``\cmark" means ``could be directly used to". And ``\xmark" means ``need to be adapted to".}
    \label{tab:ptms}%
    \begin{tabular}{lcccccccc}
    \toprule
      & \tabincell{c}{BERT\\RoBERTa} & \tabincell{c}{ZEN\\NEZHA\\ERNIE-1.0/2.0} & PanGu-$\alpha$ & CPM & CPM-2 & BART & CPT \\
    \midrule
    \multirow{3}[0]{*}{\# Params} & Base - 110M &   & 32Layers - 2.6B & Small - 110M & Base - 11B & Base - 139M & Base - 121M \\
      & Large - 340M & $\approx$ BERT & 40Layers - 13.1B & Medium - 340M & MOE - 198B & Large - 406M  & Large - 393M  \\
      &   &    & 64Layers - 207.0B & Large - 2.6B &   &   &  \\
    \midrule
    \multirow{2}[0]{*}{Arch.} & Transformer & Transformer & Transformer & Transformer & Full & Full & Unbalanced Full \\
      & Encoder & Encoder Variant & Decoder & Decoder & Transformer & Transformer & Transformer \\
    \tabincell{l}{PreTrain.\\Task} & MLM & MLM & LM & LM & Seq2Seq MLM & DAE & MLM+DAE\\
    \midrule
    Tok. & Char & Char & Word/Char & Word/Char & Word/Char & Char & Char \\
    Masking & Word & - & - & - & - & Word & Word \\
    Prediction & Char & Char & Word/Char & Word/Char & Word/Char & Char & Char \\
    \midrule
    NLU & \cmark & \cmark & \xmark & \xmark & \xmark & \xmark & \cmark \\
    NLG & \xmark & \xmark & \cmark & \cmark & \cmark & \cmark & \cmark \\
    \bottomrule
    \end{tabular}%
\end{table*}%

The main properties of CPT are as follows:
\begin{itemize}
  \itemindent 2.8em
  \item[(1)] CPT can be regarded as two separated PTMs with a shared encoder. Two specific decoders are pre-trained with MLM and DAE tasks, respectively. Each decoder can learn the specific knowledge on either NLU or NLG tasks, while the shared encoder learns the common knowledge for universal language representation.
  \item[(2)] Two separated decoders enable CPT to adapt to various downstream tasks flexibly. For example, CPT could be fine-tuned with at least five modes for classification tasks (as shown in Figure \ref{fig:finetune-classification}), which exploits the full potential of CPT. Thus, we could choose a suitable fine-tuning mode based on the attributes and characteristics of downstream tasks.
  \item[(3)] The overall architecture of CPT is an unbalance Transformer. To make the computational cost and the size of CPT comparable  with popular PTMs, such as BERT and BART, we use a novel architecture consisting of a deeper shared encoder and two shallower decoders. Especially, the shallow generation decoder greatly accelerates the inference of text generation.
\end{itemize}

We conduct experiments on various language understanding and text generation tasks, including datasets for text classification, sequence labeling, machine reading comprehension, summarization, data-to-text generation, etc. Results show that CPT could achieve competitive results with state-of-the-art on these datasets.

\section{Related Work}

\subsection{PTMs towards both NLU and NLG}

Recently, there are some efforts to combine language understanding and generation into a single pre-trained model.
UniLM~\cite{Dong19UniLM} pre-trained with an ensemble of attention masks, which allows the model to be used for both generative and classification tasks. A difference is that all parameters of UniLM are shared between generation and discrimination, whereas CPT uses two separated decoders. Thus, CPT can utilize the DAE pre-training task which is proven to be effective for NLG tasks~\cite{Lewis20BART}.

PALM~\cite{Bi20PALM} is a pre-trained model focusing on conditional generation. To force the encoder to comprehend the meaning of the given context, MLM is added to pre-train the encoder.
In contrast, CPT has an individual decoder for MLM which can avoid the negative effects brought by DAE. Therefore CPT also has good performance on NLU tasks.

More recently, ERNIE 3.0~\cite{Sun21ernie3} also uses a universal encoder and several task-specific decoders, but it adopts Transformer-XL as the backbone and its generative pre-training task is left-to-right LM with a special masked attention matrix. Different from ERNIE 3.0, CPT adopts the encoder-decoder architecture and is more suitable for sequence-to-sequence (Seq2Seq) tasks.

\subsection{Chinese PTMs}

Many attempts have been conducted to pre-train the Chinese counterparts of PTMs.

The first line of works follows BERT and uses MLM with whole word masking strategy to pre-train Transformer encoder, such as
Chinese versions of BERT and RoBERTa~\cite{Cui19cnbert},
NEZHA~\cite{Wei19nezha}, ZEN~\cite{Diao20zen}.
Some of them add special features of Chinese characters or words to further boost the performance of NLU tasks, such as ERNIE 1.0/2.0~\cite{Sun19ernie,Sun20ernie2}, ChineseBERT~\cite{Sun21Chinesebert}. However, these PTMs could not be adopted to text generation directly.

The second line of works follows GPT and uses the left-to-right LM task to pre-train a Transformer decoder, such as CPM~\cite{Zhang21cpm2} and PanGu~\cite{Huawei21Pangu}.
Although large-scale PTMs with tens of billions parameters have been released recently, the huge computation and storage cost hinders their applications.

The third line of works aims to pre-train the full Transformer encoder-decoder. CPM-2~\cite{Zhang21cpm2} follows T5~\cite{Google20T5} and adopts a Seq2Seq MLM pre-training task, which predicts the masked tokens in a Seq2Seq fashion.
Although BART~\cite{Lewis20BART} has achieved widely success on conditional text generation tasks, such as text summarization~\cite{Dou21GSum,Liu21SimCLS} and dialogue system~\cite{Lin20BART-dialogue}, it still lacks corresponding Chinese versions\footnote{Besides CPT, we also provide a Chinese BART as a byproduct.}.

Different from the above Chinese PTMs, CPT is a pre-trained unbalanced Transformer with MLM and DAE tasks, which is capable of achieving competitive results on both NLU and NLG tasks. Besides, CPT is parameter efficient compared to these large-scale models. Table~\ref{tab:ptms} compares different Chinese PTMs.

\subsection{Multi-Task Pre-Training}
Incorporating multi-task learning into pre-training has drawn increasingly attention recently. Most recent advancements attempt to improve performance by leveraging multi-task learning beyond standard pre-training~\cite{Liu2019MTDNN,Google20T5,Aghajanyan2021Muppet,Wei2021FLAN}. This line of works focuses on downstream task performance improvements by utilizing a collection of labeled datasets. However, our work is focusing on close the gap between language understanding and text generation tasks by applying multi-task learning on large scale unlabeled texts.

\section{Model Architecture}

As shown in Figure~\ref{fig:pretrain}, The architecture of CPT is a variant of the full Transformer and consists of three parts:
\begin{itemize}
\itemindent 2.8em
  \item[(1)] \textbf{Shared Encoder} (S-Enc): a Transformer encoder with fully-connected self-attention, which is designed to capture the common semantic representation for both language understanding and generation.
  \item[(2)] \textbf{Understanding Decoder} (U-Dec): a shallow Transformer encoder with fully-connected self-attention, which is designed for NLU tasks. The input of U-Dec is the output of S-Enc.
  \item[(3)] \textbf{Generation Decoder} (G-Dec): a Transformer decoder with masked self-attention, which is designed for generation tasks with auto-regressive fashion. G-Dec utilizes the output of S-Enc with cross-attention.
\end{itemize}

With the two specific decoders, CPT can be used flexibly. For example, CPT can be easily fine-tuned for NLU tasks using just S-Enc and U-Dec, and can be regarded as the standard Transformer encoder; while for NLG tasks, CPT adopts S-Enc and G-dec, and forms a Transformer encoder-decoder.
With different combinations, CPT is able to be effectively applied on various downstream tasks, which fully exploits the pre-trained parameters and obtains competitive performance.
More combinations and use cases will be discussed in \textbf{Fine-Tuning} Section.

Different from most PTMs with encoder-decoders, we exploit a deep-shallow framework for shared encoder and decoders. More specifically, we use a deeper encoder and two shallow decoders for CPT.
We assume that a shallow decoder retains the performance on text generation and reduces decoding time, which has proven to be effective for neural machine translation~\cite{Kasai21deepshallow} and spell checking~\cite{Sun21shallowGEC}.

The deep-shallow setup makes CPT more general for both understanding and generative tasks with minor parameter overheads. It also accelerates the inference of CPT for text generation as the G-Dec is a light decoder.

\section{Pre-Training}
To make CPT good at both NLU and NLG tasks, we introduce two pre-training tasks.
\begin{itemize}
\itemindent 2.8em
  \item[(1)] \textbf{Masked Language Modeling} (MLM): We pre-train the parameters of S-Enc and U-Dec with MLM~\cite{Devlin19bert,Cui19cnbert}. Given a sentence, we randomly replace some tokens with the \texttt{[MASK]} token and train S-Enc and U-Dec to predict the masked tokens. Following~\cite{Cui19cnbert}, we adopt Whole Word Masking (WWM) to replace the tokens.  Compared to randomly token masking, WWM is more suitable for inducing semantic information carried by words and spans.
  \item[(2)] \textbf{Denoising Auto-Encoding} (DAE): We pre-train the parameters of S-Enc and G-Dec by reconstructing the original document based on the corrupted input. According to the studies of BART~\cite{Lewis20BART}, we corrupted the input by two effective ways. 1) \textit{\textbf{Token Infilling}}: a Whole Word Masking (WWM) strategy with single mask replacement. First, a number of words are sampled based on the segmentation. Then, each selected word is replaced with a single \texttt{[MASK]} token, regardless of how many tokens it consists; and 2) \textit{\textbf{Sentence Permutation}}: sentences are extracted from a document based on punctuation, and shuffled in a random order.
\end{itemize}

In practice, We first use a Chinese Word Segmentation (CWS) tool to split the sentences into words. Then, we select 15\% of the words and mask the corresponding characters. For the masked characters, we follow the setup of BERT to (1)  replace 80\% of them with a special \texttt{[MASK]} token, (2) replace 10\% of them by random tokens, (3) keep the rest 10\% of them unchanged.

Finally, we train CPT with two pre-training tasks under a multi-task learning framework. Thus, CPT can learn for both understanding and generation, and can easily deal with downstream NLU or NLG tasks.

\section{Fine-Tuning}
\label{sec:finetune}

PTMs are usually fine-tuned in only few ways for a given downstream task. For example, for sentence-level classification, we fine-tune BERT by taking the top-layer output of \texttt{[CLS]} token as the representation of the whole sentence, while fine-tune GPT by using the representation of the last token of the sequence.

Thanks to the separated understanding and generation decoders, CPT can be fine-tuned in multiple patterns. For a given downstream task, one could choose the most suitable way to fully stimulate the potential of CPT to achieve competitive results.

\subsection{Fine-Tuning for Sentence-Level Classification}

\begin{figure*}[t!]
    \centering
    \begin{minipage}[t]{0.31\textwidth}
        \centering
        \includegraphics[page=1,width=0.5\textwidth]{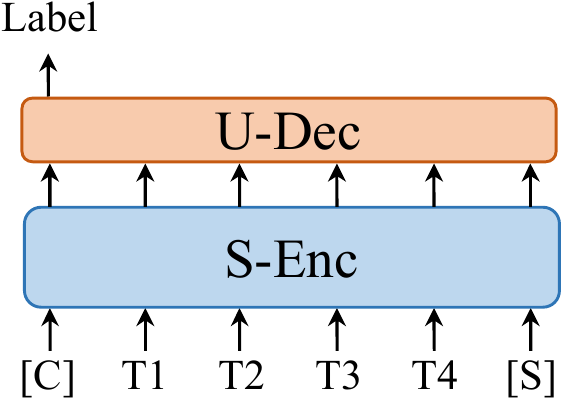}
    \end{minipage}
    \begin{minipage}[t]{0.31\textwidth}
        \centering
      \includegraphics[page=2,width=\textwidth]{cpt-graphs-crop.pdf}
    \end{minipage}
    \hspace{0.05\textwidth}
    \begin{minipage}[t]{0.31\textwidth}
        \centering
        \includegraphics[page=3,width=\textwidth]{cpt-graphs-crop.pdf}
    \end{minipage}
    \\
    \begin{minipage}[t]{0.31\textwidth}
      \centering
      \caption*{(a) CPT$_u$}
      \label{fig:du-only}
  \end{minipage}
    \begin{minipage}[t]{0.31\textwidth}
      \centering
      \caption*{(b) CPT$_g$}
      \label{fig:dg-only}
    \end{minipage}
    \hspace{0.05\textwidth}
    \begin{minipage}[t]{0.31\textwidth}
        \centering
        \caption*{(c) CPT$_{ug}$}
        \label{fig:both-dudg}
    \end{minipage}
    \\
    \begin{minipage}[t]{0.31\textwidth}
        \centering
        \includegraphics[page=5,width=0.7\textwidth]{cpt-graphs-crop.pdf}
    \end{minipage}
    \begin{minipage}[t]{0.31\textwidth}
        \centering
        \includegraphics[page=4,width=0.9\textwidth]{cpt-graphs-crop.pdf}
    \end{minipage}
    \\
    \begin{minipage}[t]{0.31\textwidth}
      \centering
      \caption*{(d) CPT$_{u+p}$}
      \label{fig:prompt-du}
      \end{minipage}
    \begin{minipage}[t]{0.31\textwidth}
      \centering
      \caption*{(e) CPT$_{g+p}$}
      \label{fig:prompt-dg}
    \end{minipage}
\caption{Five ways to fine-tune CPT for text classification. ``T1-4" and ``P1-2" refer to text input $\mathbf{x}$ and prompt tokens, respectively. $\mathcal{V}(y)$ is the mapping function that maps the language model predictions to the label. \texttt{[C]} and \texttt{[S]} are abbreviations for \texttt{[CLS]} and \texttt{[SEP]}, respectively.}
\label{fig:finetune-classification}
\end{figure*}

When incorporating external classifiers, CPT have three fine-tuning modes for sequence-level classification (As shown in Figure~\ref{fig:finetune-classification} (a),(b) and (c)).

\begin{itemize}
\itemindent 2.8em
  \item[(1)] CPT$_{u}$: a BERT-style mode. The sentence representation is from U-Dec module only, which is usually the first state of \texttt{[CLS]} token.
  \item[(2)] CPT$_{g}$: a BART-style mode.  The same input is fed into the S-Enc and G-Dec, and the representation from the final output token \texttt{[SEP]} from G-Dec is used.
  \item[(3)] CPT$_{ug}$: The same input is fed into the S-Enc and G-Dec, and the final representation is the concatenation of the first output of U-Dec and the final output of G-Dec.
\end{itemize}

Recently, a powerful and attractive framework, prompt-based learning~\cite{Schick21pet,Gao20LMBFF,Liu21prompt-survey}, is also able to boost the performance of PTMs. By defining prompting templates and reformulating the classification tasks into a generative fashion, the framework utilizes PTMs to generate words corresponding to task labels. The generative patterns are so close to the pre-training tasks of PTMs that they have the ability of few-shot or even zero-shot learning.

The prompt-based methods could also be applied on CPT with more flexibly fashions since CPT has two decoders. As shown in Figure~\ref{fig:finetune-classification} (d) and (e), we construct prompts and convert the task into an generation task with CPT by the following two modes:
\begin{itemize}
\itemindent 2.8em
  \item[(1)] CPT$_{u+p}$: A MLM task. We manually construct an input template and assign a word to each task label. CPT is fine-tuned to predict the word at the masked positions, which will be mapped to the task labels. Since a word may be tokenized into multiple character tokens, the predicted distributions at masked positions are averaged to get the predicted distribution of labels.
  \item[(2)] CPT$_{g+p}$: Conditional text generation. We encode the input text with S-Enc and train CPT to generate prompt text initialized with corresponding labels by teacher forcing. For inference, we first construct the prompt text for each label. Then, the perplexity of each prompt text is calculated. Finally, the prediction is assign to the label with the highest corresponding perplexity.
\end{itemize}

\subsection{Fine-Tuning for Sequence Labeling}

\begin{figure*}[ht!]
    \centering
    \begin{minipage}[t]{0.48\textwidth}
        \centering
        \includegraphics[page=6, width=0.4\textwidth]{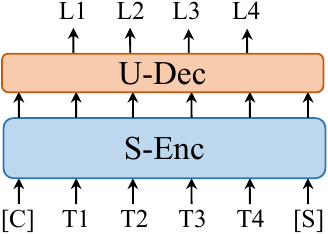}
    \end{minipage}
    \hspace{0.02\textwidth}
    \begin{minipage}[t]{0.48\textwidth}
      \centering
      \includegraphics[page=7, width=0.8\textwidth]{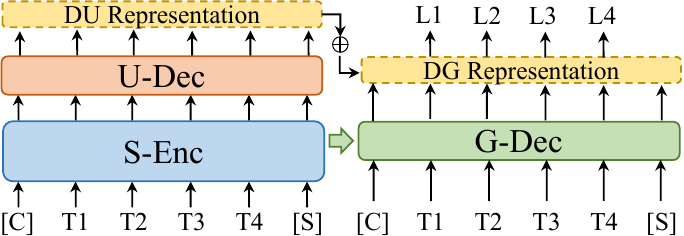}
    \end{minipage}
    \begin{minipage}[t]{0.48\textwidth}
      \centering
      \caption*{(a) CPT$_u$}
      \label{fig:seq-label-u}
      \end{minipage}
      \hspace{0.02\textwidth}
    \begin{minipage}[t]{0.48\textwidth}
      \centering
      \caption*{(b) CPT$_{ug}$}
      \label{fig:seq-label-ug}
    \end{minipage}
\caption{Two examples of fine-tuning CPT for sequence labeling. ``T1-4" and ``L1-4" refer to  text input $\mathbf{x}$ and token labels, respectively.}
\label{fig:finetune-seq-label}
\end{figure*}

For sequence labeling, each token needs a representation for token-level classification. Similar to sequence-level classification, we leverage PTMs to obtain high quality token representations and then put the representations to a trainable classifier to assign labels for these tokens. Thus, similar to sentence-level classification, we can fine-tune CPT for sequence labeling as CPT$_{u}$, CPT$_{g}$ and CPT$_{ug}$, using (1) U-Dec only, (2)  G-Dec only, or (3) both U-Dec and G-Dec.  Figure~\ref{fig:finetune-seq-label} shows two examples for sequence labeling.

\subsection{Fine-Tuning for Machine Reading Comprehension}

\begin{figure*}[ht!]
  \centering
  \begin{minipage}[t]{0.48\textwidth}
    \centering
    \includegraphics[page=8,width=0.4\textwidth]{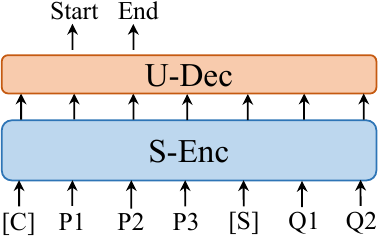}
  \end{minipage}
  \hspace{0.02\textwidth}
  \begin{minipage}[t]{0.48\textwidth}
    \centering
    \includegraphics[page=10,width=0.8\textwidth]{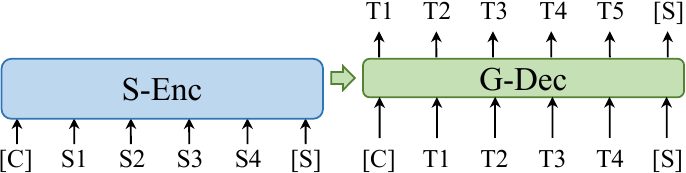}
  \end{minipage}
  \\
  \begin{minipage}[t]{0.48\textwidth}
      \centering
      \caption{Two examples of fine-tuning CPT for Machine Reading Comprehension. ``P1-3", ``Q1-2" refer to passages and questions, respectively.}
      \label{fig:mrc-u}
  \end{minipage}
  \hspace{0.02\textwidth}
  \begin{minipage}[t]{0.48\textwidth}
      \centering
      \caption{Example of fine-tuning CPT$_g$ for Conditional Generation. ``S1-4'' and ``T1-4" refer to input and target sequences, respectively.}
      \label{fig:finetune-summary}
  \end{minipage}
\end{figure*}

Machine Reading Comprehension requires the model to predict an answer span shown in the passage for a given question. A typical fine-tuning pattern is to train PTMs to predict the start and end positions of the span in the passage. The prediction is based on the tokens of the passage. Thus, CPT$_{u}$, CPT$_{g}$ and CPT$_{ug}$ can be fine-tuned, similar to sequence-labeling. Figure~\ref{fig:mrc-u} shows the example of CPT$_{u}$.

\subsection{Fine-Tuning for Conditional Generation}

Apart from NLU tasks, CPT can do text generation efficiently. As shown in Figure~\ref{fig:finetune-summary}, we simply fine-tune CPT$_g$ with S-Enc and G-Dec modules on text generation tasks, similar to the usage of other auto-regressive PTMs~\cite{Lewis20BART}.

\section{Experiments}

\subsection{Pre-Training Setups}
\label{sec:pretrain}
We implement two versions of CPT, namely, \textit{base} and \textit{large}, respectively consisting of 14/28 Transformer layers with 10/20 layers for shared encoder and 2/4 layers for each task specific decoder. And the hidden units and attention heads per layer for base and large versions are 768/1,024 and 12/16, respectively.
The total number of layers activated for a given task is always equal to 12/24, which makes our model comparable with base/large-size of BERT and its variants (RoBERTa, ERNIE 1.0/2.0, etc).

We train our models on the open source large-scale raw text, Chinese Wikipedia and a part of WuDaoCorpus. The training data contains 200GB cleaned text ranges from  different domains. We use Jieba to segment Chinese words for Whole Word Masking and use WordPiece tokenizer inherited from BERT to split input text into tokens. We use Adam to train the models for 500k steps, with the batch size of 2048, the learning rate of 1e-4, $\beta_1 = 0.9 $, $\beta_2 = 0.98$, weight decay of 0.01. We warmup the learning rate for first 10,000 steps then do linear decay. In addition, a \textbf{Chinese BART} is pre-trained with the same corpora, tokenization and hyper-parameters as a baseline.

\subsection{Evaluation Tasks}
To evaluate the effectiveness of our model, we conduct experiments on various NLP datasets across different understanding and generation tasks, with details illustrated below.

\paragraph{Classification} We evaluate the model on the Chinese Language Understanding Evaluation Benchmark (CLUE)~\cite{CLUEbenchmark}, which contains text classification \textbf{TNEWS}, \textbf{IFLYTEK}, natural language inference (NLI), \textbf{OCNLI}, sentence pair matching (SPM) \textbf{AFQMC}, and coreference resolution (CoRE) CLUEWSC 2020 (\textbf{WSC.}) key word recognition (KwRE) \textbf{CSL}. We conduct data augmentation \textbf{CSL} as \cite{Zhang20ambert} performed, and evaluate \textbf{TNEWS} on version 1.1 test set. Accuracy is used for these datasets.

\paragraph{Sequence Labeling} We evaluate our model on Chinese word segmentation (CWS) and named entity recognition (NER), which are two representative sequence labeling tasks. We use two datasets from \textbf{SIGHAN2005}~\cite{Emerson05sighan} for CWS, which are \textbf{MSR}, \textbf{PKU}. And for NER, \textbf{MSRA}~\cite{Levow06msraner}, \textbf{OntoNotes}\footnote{https://catalog.ldc.upenn.edu/LDC2011T03} are used. We use the same dataset preprocessing and split methods as in previous work~\cite{Li21seqlabel-ee,Li2020flat,Qiu20multicws}. And F1 scores are reported.

\paragraph{MRC} Span based machine reading comprehension (MRC) dataset CMRC 2018 (\textbf{CMRC})~\cite{Cui19cmrc18} and Traditional Chinese MRC dataset \textbf{DRCD}~\cite{Shao18drcd} are used. We follow the data processing in \cite{Cui19cnbert,Cui20macbert} and transform the text from \textbf{DRCD} is transformed to Simplified Chinese. The Exact Match (EM) scores are reported.

\paragraph{Text Generation} We use two abstractive summarization datasets, \textbf{LCSTS}~\cite{Hu15lcsts} and \textbf{CSL}\footnote{https://github.com/CLUEbenchmark/CLGE}, and a data-to-text generation dataset, \textbf{ADGEN}~\cite{Shao19adgen} to evaluate the text generation ability of our model. Among them, \textbf{LCSTS} is a large corpus of Chinese short text summarization dataset constructed from Sina Weibo, consisting of 2 million real Chinese short texts with short summaries. And \textbf{CSL} is an academic domain text summarization dataset, constructed from abstract and titles from publications in computer science domain. And \textbf{ADGEN} is a data-to-text dataset that requires models to generate long text for advertisement based on some keywords.
And we evaluate PTMs on test sets of \textbf{LCSTS} and \textbf{ADGEN} and the development set of \textbf{CSL}.
The character-level Rouge-L is used to evaluate the summarization results. For \textbf{ADGEN}, we follow \cite{Zhang21cpm2} to use BLEU-4.

\subsection{Compared PTMs}
We compare CPT with a series of state-of-the-art PTMs for either natural language understanding or text generation. The details are as follows.
\paragraph{PTMs for NLU}
PTMs with the Transformer Encoder structure and pre-trained with MLM usually perform well in NLU tasks, such as the Chinese versions of BERT and RoBERTa~\cite{Cui19cnbert}, NEZHA~\cite{Wei19nezha}, ERNIE 2.0~\cite{Sun20ernie2}, MacBERT~\cite{Cui20macbert}.
Unless otherwise specified, we use BERT and RoBERTa to refer to \texttt{BERT-wwm-ext} and \texttt{RoBERTa-wwm-ext}, respectively.

\paragraph{PTMs for NLG}
For text generation, we compare CPT with generative Transformers ranging from normal size to large scale, including BART~\cite{Lewis20BART}, mBART~\cite{Liu20mbart}, mT5~\cite{Xue21mT5}, CPM-2~\cite{Zhang21cpm2}, and models with pre-trained encoders.
BART is a sequence-to-sequence model pre-trained with DAE task.
Due to the missing of Chinese version, we train a Chinese BART as mentioned in Section~\ref{sec:pretrain}.
mBART is a multilingual variant of BART.
And mT5 is a multilingual variant of T5 pre-trained on over 101 languages, including Chinese.
CPM-2 is a large-scale encoder-decoder model with 11 billion parameters,   pre-trained in multiple stages with large-scale Chinese and bilingual data.
We also report generative models adopted from Transformer encoders such as RoBERTa and ERNIE 2.0 that follow the generation style of UniLM~\cite{Dong19UniLM}, to further evaluate the effectiveness generative pre-training.

\subsection{Main Results}

To fully release the potential of our model, we fine-tune CPT for NLU tasks in different ways as mentioned in \textbf{Fine-Tuning} Section, denoted as CPT$_u$, CPT$_g$ and CPT$_{ug}$, CPT$_{u+p}$ and CPT$_{g+p}$, respectively.
We use (B) and (L) to distinguish base and large version of PTMs, respectively.

\paragraph{Classification}
Table~\ref{tab:classification-dev} shows the development set results of CLUE Benchmark of different fine-tuning modes. As a result, CPT$_u$ (B) achieves a 74.6 on average, surpassing other baselines and fine-tuning patterns on base version of CPT. Besides, CPT$_{ug}$ (L) obtains a averaged accuracy 76.2, which is better than RoBERTa (L) by a large margin. Therefore, we choose CPT$_u$ (B) and CPT$_{ug}$ (L) as the most suitable fine-tuning patterns to do the classification. We find that the best fine-tuning modes are different between base and large models. We believe the difference is brought by the scale of the parameters. For base model, the G-Dec is too shallow to transfer for NLU tasks, which makes CPT$_ug$ could not beat the CPT$_u$. And the G-Dec in large version of CPT has more parameters and layers, which makes the decoder easy to transfer.

\begin{table*}[ht!]
  \centering
  \fontsize{8.3}{10}\selectfont
  \setlength{\tabcolsep}{.5mm}
  \caption{Accuracy results on dev set of CLUE Benchmark. We fine-tune CPT with five different ways as shown in Figure~\ref{fig:finetune-classification}. (B) and (L) refer to base-size and large-size of PTMs, respectively.}
  \label{tab:classification-dev}%
  \begin{tabular}{lccccccc}
    \toprule
    \textbf{Models} & \textbf{TNEWS} & \textbf{IFLYTEK} & \textbf{OCNLI} & \textbf{AFQMC} & \textbf{CSL} & \textbf{WSC} & \textbf{AVG} \\
  \midrule
  BERT (B) & 56.8 & 58.9 & 75.4 & 72.0 & 82.3 & 83.2 & 71.4 \\
  RoBERTa (B) & 57.5 & 59.4 & 76.5 & 74.4 & 86.1 & 88.8 & 73.8 \\
  BART (B) & 57.2 & 60.0 & 76.1 & 73.0 & 85.8 & 79.6 & 71.9 \\
  \rowcolor{gray!15}
  CPT$_u$ (B) & \textbf{58.4} & 60.5 & 76.4 & \textbf{75.1} & 86.1 & \textbf{91.1} & \textbf{74.6} \\
  CPT$_g$ (B) & 57.3 & 60.4 & 76.3 & 71.4 & 86.4 & 87.2 & 73.2 \\
  CPT$_{ug}$ (B) & 57.4 & \textbf{61.9} & \textbf{76.8} & 70.6 & 86.3 & 89.8 & 73.8 \\
  CPT$_{g+p}$ (B) & 54.9 & 25.4 & 76.6 & 73.7 & \textbf{86.9} & 79.9 & 66.2 \\
  CPT$_{u+p}$ (B) & \textbf{58.4} & 61.6 & 76.6 & \textbf{75.1} & \textbf{86.9} & 79.9 & 73.1 \\
  \midrule
  RoBERTa (L) & 58.3 & 61.7 & 78.5 & 75.4 & 86.3 & 89.5 & 75.0 \\
  BART (L) & \textbf{59.2} & 62.1 & 79.7 & 75.7 & \textbf{87.3} & 90.1 & 75.7 \\
  CPT$_u$ (L) & 58.8 & 61.8 & 79.5 & \textbf{75.9} & 86.5 & 92.1 & 75.8 \\
  CPT$_g$ (L) & 59.1 & 61.7 & \textbf{79.9} & 75.8 & 86.9 & 91.8 & 75.9 \\
  \rowcolor{gray!15}
  CPT$_{ug}$ (L) & \textbf{59.2} & \textbf{62.4} & 79.8 & 75.8 & 86.6 & \textbf{93.4} & \textbf{76.2} \\
  CPT$_{g+p}$ (L) & 54.5 & 29.2 & 79.8 & 75.4 & 87.1 & 89.5 & 69.2 \\
  CPT$_{u+p}$ (L) & 59.0 & 61.2 & 79.6 & 75.4 & \textbf{87.3} & 87.8 & 75.1 \\
  \bottomrule
  \end{tabular}%
\end{table*}%

\begin{table*}[ht!]
  \centering
  \fontsize{8.3}{10}\selectfont
  \setlength{\tabcolsep}{.5mm}
  \caption{Results on CLUE benchmarks. For all tasks we report accuracy on test sets.}
  \label{tab:exp-clue}
  \begin{tabular}{lccccccc}
      \toprule
      \textbf{Models} & \textbf{TNEWS} & \textbf{IFLYTEK} & \textbf{OCNLI} & \textbf{AFQMC} & \textbf{CSL} & \textbf{WSC} & \textbf{AVG} \\
      \midrule
      BERT (B) & 58.6 & 59.4 & 73.2 & 74.1 & 84.2 & 74.5 & 70.7 \\
      RoBERTa (B) & \textbf{59.5} & 60.3 & \textbf{73.9} & 74.0 & 84.7 & 76.9 & 71.5 \\
      BART (B) & 58.5 & \textbf{60.7} & 72.1 & 74.0 & 85.4 & 67.6 & 69.7 \\
      \rowcolor{gray!15}
      CPT$_u$ (B) & 59.2 & 60.5 & 73.4 & \textbf{74.4} & \textbf{85.5} & \textbf{81.4} & \textbf{72.4} \\
      \midrule
      RoBERTa (L) & 58.9 & \textbf{63.0} & 76.4 & \textbf{76.6} & 82.1 & 74.6 & 71.9 \\
      BART (L) & 58.6 & 62.7 & 78.1 & 74.3 & \textbf{86.7} & 82.1 & 73.7 \\
      \rowcolor{gray!15}
      CPT$_{ug}$ (L) & \textbf{59.2} & 62.4 & \textbf{78.4} & 75.0 & 85.5 & \textbf{86.2} & \textbf{74.5} \\
      \bottomrule
      \end{tabular}%
\end{table*}

For prompt-based fine-tuning (Table~\ref{tab:classification-dev}), we find that directly fine-tuning without prompt works well on some datasets, with the small gaps between CPT$_u$, CPT$_g$ and CPT$_{ug}$.
Moreover, CPT$_{u+p}$ achieves good results on some datasets that even outperform methods without prompt tuning. However, the accuracy of prompt-base methods on other datasets drops a lot. As there are many factors that affect prompt tuning performance including prompt design, choices of words for labels, etc. Manually designed prompts may be suboptimal.
Besides, we find that CPT$_{g+p}$ degenerates obviously on TNEWS and IFLYTEK. Both datasets have more than 3 classes, which contains 15 and 112 labels, respectively. Moreover, these labels are hard to represented by a single character. In practice we assign words with up to 7 characters to a label. We presume that the large number of labels and the multi-token issue hinders CPT$_{g+p}$ to generate correctly.

Table~\ref{tab:exp-clue} reports the performance of CPT on classification tasks and the comparison with previous representative Chinese PTMs. We report accuracy on the test sets of these datasets. Among the fine-tuned CPTs, we choose base version CPT$_u$ and large version CPT$_{ug}$ as they obtain the best results on development sets. Base size CPT consistently outperforms BERT, RoBERTa and ERNIE. Moreover, large size CPT achieves a 74.5 averaged score, outperforming RoBERTa (L) with a large margin.
We find that generative PTMs, such as BART, also have the ability to handle discrimination tasks (see Table~\ref{tab:classification-dev} Table~\ref{tab:exp-clue}). However, their performance is suboptimal compared with the CPT. As the uni-directional layers of generative models could hurt the performance of NLU tasks.

\paragraph{Sequence Labeling}
The CPT is fine-tuned as CPT$_u$, CPT$_g$ and CPT$_{ug}$ and evaluated on development sets.
We find that CPT$_u$ constantly obtains the best development results. We conjecture that CWS and NER have more dependency on local syntax than complex semantics used for text generation. Thus, CPT$_u$ is more suitable for CWS and NER with its bidirectional fully connected self-attention.
As a result, we report the test set results of CPT$_u$ to compare with other PTMs.

\begin{table}[th!]
  \centering
  \footnotesize
  \caption{Results on sequence labeling datasets. The F1 scores on test sets are reported. Models with $^*$ indicate the results are from~\cite{Sun20ernie2}.}
  \label{tab:exp-seq-label}
  \begin{tabular}{lcccc}
  \toprule
    & \multicolumn{2}{c}{\textbf{CWS}} & \multicolumn{2}{c}{\textbf{NER}} \\
    \cmidrule(lr{1em}){2-3} \cmidrule(lr{1em}){4-5}
    & \textbf{MSR} & \textbf{PKU} & \textbf{MSRA} & \textbf{OntoNotes} \\
  \midrule
  BERT (B) & 98.24 & 96.50 & 95.13 & 81.73 \\
  ERNIE 2.0$^*$ (B) & - & - & 93.80 & - \\
  RoBERTa (B) & 98.14 & 96.15 & 95.23 & 81.52 \\
  \rowcolor{gray!15}
  CPT$_u$ (B) & \textbf{98.29} & \textbf{96.58} & \textbf{95.78} & \textbf{82.08} \\
  \midrule
  ERNIE 2.0$^*$ (L) & - & - & 95.00 & - \\
  RoBERTa (L) & 98.42 & 96.37 & 95.20 & 81.78 \\
  \rowcolor{gray!15}
  CPT$_u$ (L) & \textbf{98.51} & \textbf{96.70} & \textbf{96.20} & \textbf{83.08} \\
  \bottomrule
  \end{tabular}%
\end{table}

We compare our model with other state-of-the-art methods on sequence labeling datasets. As shown in Table~\ref{tab:exp-seq-label}, CPT$_u$ (L) achieves the highest performance and exceed the BERT (L), RoBERTa (L) and ERNIE (L) on all sequence labeling tasks, both CWS and NER. And CPT$_u$ (B) obtains a comparable results, surpassing base versions of BERT and RoBERTa.

Note that CPT$_ug$ outperforms the CPT$_u$ in the large size while surpassed by CPT$_u$ in the base version. We believe that it is the large discrepancy between pre-training and fine-tuning tasks, which makes the G-Dec trained by the DAE task hard to be transferred to classification. G-Dec is harder to be fine-tuned than understanding decoder (U-Dec), especially in the base model where G-Dec is very shallow. And it also explains that the performance gap between CPT$_u$ and CPT$_g$ in the base version is larger than the large size.  

\begin{table}[ht!]
  \centering\footnotesize
  \caption{Results on MRC datasets. Exact Match (EM) scores are reported. Models with $^*$ indicate the results from the corresponding work.}
  \label{tab:exp-mrc}
  \begin{tabular}{lccc}
  \toprule
        & \textbf{CMRC 2018} & \multicolumn{2}{c}{\textbf{DRCD}} \\
        \cmidrule(lr{1em}){2-2} \cmidrule(lr{1em}){3-4}
        & \textbf{Dev} & \textbf{Dev} & \textbf{Test} \\
  \midrule
      RoBERTa (B) & 67.9 & 85.9 & 85.2 \\
      MacBERT$^*$ (B) & 68.2 & \textbf{89.2} & 88.7 \\
      ERNIE 2.0$^*$ (B) & \textbf{69.1} & 88.5 & 88.0 \\
      NEZHA$^*$ (B) & 67.8 & - & - \\
      \rowcolor{gray!15}
      CPT$_u$ (B) & 68.8 & 89.0 & \textbf{89.0} \\
      \midrule
      RoBERTa (L) & 70.6 & 89.1 & 88.9 \\
      MacBERT$^*$ (L) & 70.1 & 90.8 & 90.9 \\
      ERNIE 2.0$^*$ (L) & 71.5 & 89.7 & 89.0 \\
      NEZHA$^*$ (L) & 68.1 & - & - \\
      \rowcolor{gray!15}
      CPT$_u$ (L) & \textbf{72.3} & \textbf{91.0} &  \textbf{91.1} \\
  \bottomrule
  \end{tabular}%
\end{table}

\paragraph{MRC}

Table~\ref{tab:exp-mrc} shows the experimental results on MRC tasks, which also indicates the effectiveness of CPT. We report the Exact Match (EM) score on CMRC dev set, DRCD dev and test sets. We try and evaluate  CPT$_u$,  CPT$_u$ and  CPT$_u$ on the development sets of these datasets and choose the pattern that acquires the best results to report. As a conclusion, CPT$_u$ obtains comparable or higher results compared to previous systems that are widely used, such as RoBERTa, MacBERT, ERNIE and NEZHA. Moreover, CPT$_u$ consistently outperforms other strong baselines by a large margin, with 72.3 EM score on the CMRC development set and 91.1 EM on the DRCD test set.

\paragraph{Text Generation}
\begin{table*}[th!]
    \centering\footnotesize
    \caption{Results on text generation datasets. The small(base) version of mT5 has almost the same parameters as the base(large) version of other PTMs. CPM-2 has a much larger number of parameters than other large size PTMs. Models with $^*$ and $^\dagger$ indicate the results are from \cite{Sun21ernie3} and \cite{Zhang21cpm2}, respectively.}
    \label{tab:exp-sum}
    \begin{tabular}{lccc}
    \toprule
    \textbf{Models} &  \textbf{LCSTS} & \textbf{CSL} & \textbf{ADGEN} \\
    & (Rouge-L) & (Rouge-L) & (BLEU-4)\\
    \midrule
    mT5 (S) &  33.5 & 56.7 & \textbf{10.2} \\
    BART (B) & 37.8 & 62.1 & 9.9 \\
    \rowcolor{gray!15}
    CPT$_g$ (B) & \textbf{38.2} & \textbf{63.0} & 9.8 \\
    \midrule
    CPM-2$^\dagger$ & 35.9 & - & 10.6 \\
    mBART (L) & 37.8 & 55.2 & 8.5 \\
    mT5 (B) &  36.5 & 61.8 & - \\
    ERNIE 2.0$^*$ (L) & 41.4 & - & - \\
    RoBERTa$^*$ (L)  & 41.0 & - & - \\
    BART (L) &  40.6 & \textbf{64.2} & 10.0 \\
    \rowcolor{gray!15}
    CPT$_g$ (L)  & \textbf{42.0} & 63.7 & \textbf{10.7} \\
    \bottomrule
    \end{tabular}%
\end{table*}

Table~\ref{tab:exp-sum} compares the performance of our model on generation datasets with other strong methods. The character-level Rouge-L is used to evaluate the summarization results. For ADGEN, we follow \cite{Zhang21cpm2} to use BLEU-4.

\begin{figure*}[th!]
  \centering
  \begin{minipage}[t]{0.48\textwidth}
    \centering
    \begin{tikzpicture}[scale=0.8]
      \begin{axis}[
        ylabel near ticks,
        ylabel={\# Tokens per Second},
        symbolic x coords={ADGEN,LCSTS,CSL},
        enlargelimits=0.25,
        legend style={at={(0.25,.96)}, anchor=north,legend columns=-1},
        yticklabels={,,},
        xtick=data,
        ybar=10pt,
        nodes near coords,
        nodes near coords align={vertical},
        nodes near coords style={/pgf/number format/.cd,fixed zerofill,precision=0}
        ]
        
        \addplot coordinates {
          (ADGEN,220.0553658)
          (CSL,242.5794589)
          (LCSTS,267.7645788)
          };
          
          \addplot coordinates {
            (ADGEN,311.8950844)
            (CSL,366.0039761)
            (LCSTS,395.3021104)
            };
            \legend{BART, CPT}
          \end{axis}
    \end{tikzpicture}
  \end{minipage}
  \begin{minipage}[t]{0.48\textwidth}
    \centering
      \begin{tikzpicture}[scale=0.8]
        \begin{axis}[
        ylabel={},
        symbolic x coords={ADGEN,LCSTS,CSL},
        enlargelimits=0.25,
        legend style={at={(0.25,.96)}, anchor=north,legend columns=-1},
        yticklabels={,,},
        xtick=data,
        ybar=10pt,
        nodes near coords,
        nodes near coords align={vertical},
        nodes near coords style={/pgf/number format/.cd,fixed zerofill,precision=0}
        ]

        \addplot coordinates {
          (ADGEN,156.8891777)
          (CSL,161.3698165)
          (LCSTS,162.339688)
          };

        \addplot coordinates {
          (ADGEN,255.7104482)
          (CSL,253.8126362)
          (LCSTS,277.8311426)
        };

        \legend{BART, CPT}
      \end{axis}
      \end{tikzpicture}
  \end{minipage}

  \caption{Inference throughput for BART and CPT. It is measured on the same parts of datasets that the models are evaluated. The beam size is 4 and the batch size is 8.}
  \label{fig:speed}
\end{figure*}
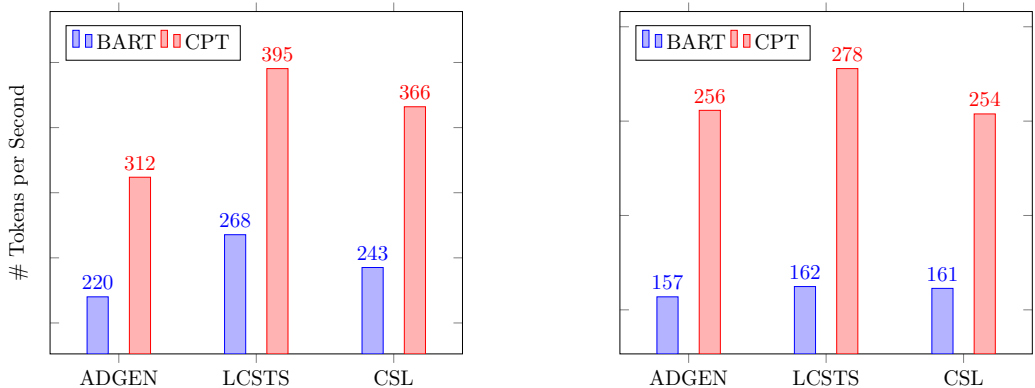

As a conclusion, CPT$_g$ achieves competitive performance on text generation compared with other methods, such as mT5, CPM-2, BART. In addition, compared with other pre-trained encoders (RoBERTa and ERNIE 2.0), CPT$_g$  improves the generation score with the NLG enhanced pre-training. When compared with pre-trained mT5 and CPM-2, CPT$_g$ acquires better results on both base and large versions. We assume the difference of pre-training tasks that lead to the performance gaps. Both mT5 and CPM-2 exploit a T5 style masked span generation as their pre-training task, while CPT is pre-trained with DAE, which shows the effectiveness of DAE for text generation pre-training.

\begin{CJK*}{UTF8}{gbsn}
  \begin{table}[th!]
    \centering\tiny
    \caption{Summary examples generated by BART (L) and CPT (L) given input text on \textbf{LCSTS}.}
    \label{tab:case-stady-summary}
    \begin{tabular}{p{0.08\textwidth}p{0.88\textwidth}}
      \hline\hline
      \multirow{2}{*}{\textbf{Input}}
       & 今日，刘胜义在2013腾讯智慧峰会上指出，在移动化时代，数字媒体、消费行为、数字营销都需要重新定义。并且移动化媒体应具备三个特征：从实时媒体发展成全天候媒体；从大众媒体发展到智能媒体阶段；从资讯媒体发展到生活类型的媒体。\\
       & Today, in the 2013 Tencent Wisdom Summit, Shengyi Liu pointed out that in the mobile era, digital media, consumer behavior, and digital marketing all need to be redefined. And mobile media should have three characteristics: real-time media develope to 24-hour media; mass media develope to smart media; information and news media develope to life media. \\
      \midrule
      \multirow{2}{*}{\textbf{Reference}} & 腾讯刘胜义：移动化引发媒体及营销体系变革 \\
        & Shengyi Liu from Tencent: Mobile process leads to the changes in media and marketing systems. \\
      \midrule
      \multirow{2}{*}{\textbf{BART (L)}} & 刘胜义：移动化时代数字媒体需重新定义 \\
        & Shengyi Liu: Digital media need to be redefined in the mobile era. \\
      \midrule
      \multirow{2}{*}{\textbf{CPT$_g$ (L)}} & 腾讯总裁刘胜义：移动化时代数字媒体需重新定义 \\
        & Tencent President Shengyi Liu: Digital media need to be redefined in the mobile era. \\
      \hline\hline
      \multirow{2}{*}{\textbf{Input}} 
        & 近年来，逢雨必涝、逢涝必瘫，几成我国城市通病。上周，中国青年报对全国31个省（区、市）5375人进行的调查显示，91.6\%的人关注所在城市的排水问题；84.7\%的受访者赞同，城市现代化更表现在地面之下，应加大地下民生工程建设投入。 \\
        & In recent years, flooding and paralysis in floods have become a common problem in Chinese cities. Last week, the China Youth Daily conducted a survey of 5375 people in 31 provinces (regions and cities) across the country. It shows that 91.6\% of people are concerned about the drainage problems in their cities; 84.7\% of the interviewees agree that urban modernization is shown under the ground, and the government should increase investment in the construction of underground livelihood projects. \\
      \midrule
      \multirow{2}{*}{\textbf{Reference}} & 84.7\%受访者期待国家加大地下民生工程投入 \\
        & 84.7\% of respondents expect the country to increase investment in underground livelihood projects. \\
      \midrule
      \multirow{2}{*}{\textbf{BART (L)}} & 84.7\%受访者赞同加大地下民生工程建设投入 \\
        & 84.7\% of respondents agree to increase investment in the construction of underground livelihood projects. \\
      \midrule
      \multirow{2}{*}{\textbf{CPT$_g$ (L)}} & 超八成受访者赞同加大地下民生工程投入 \\
      & Over 80\% of respondents agree to increase investment in the underground livelihood projects. \\
      \hline\hline
    \end{tabular}
  \end{table}
\end{CJK*}
\begin{CJK*}{UTF8}{gbsn}
  \begin{table}[t!]
    \centering\tiny
    \caption{Text examples generated by BART (L) and CPT (L) given keywords on \textbf{ADGEN}.}
    \label{tab:case-stady-text}
    \begin{tabular}{p{0.08\textwidth}p{0.88\textwidth}}
      \hline\hline
      \multirow{2}{*}{\textbf{Input}}
       & [类型, 上衣], [版型, 宽松], [颜色, 蓝色], [风格, 简约], [风格, 清新], [衣样式, 衬衫], [衣领型, 翻领], [衣长, 中长款], [衣门襟, 单排扣] \\
       & [Type, Top], [Fit, Loose], [Color, Blue], [Style, Simple], [Style, Fresh], [Clothing Style, Shirt], [Collar Type, Lapel], [Cloth Length , Mid-length], [Clothes placket, Single-Breasted] \\
      \midrule
      \multirow{2}{*}{\textbf{Reference}} 
        & 很适合学生穿的一件\textbf{衬衫}，\textbf{蓝色}调是属于比较小\textbf{清新}的色调，而且还能衬托出女性的完美好气色哦，让你穿上之后瞬间有个好心情； \textbf{简约}的小\textbf{翻领}设计加以精致的\textbf{单排扣}点缀，颇有一番正式的感觉，结合上\textbf{宽松}以及\textbf{中长款}的设计，从而打造出女性的率性bf风范。 \\
        & A \textbf{shirt} that is very suitable for students. The \textbf{blue} tone is a relatively \textbf{fresh} hue, which can bring out the perfect look of women, making you have a good mood instantly when wearing it; the \textbf{simple} small design of \textbf{lapel} is combined with \textbf{single-breasted} embellishment, which shows quite a formal feel. And when combined with \textbf{loose} and \textbf{mid-length} design, it creates a feminine and casual bf style. \\
      \midrule
      \multirow{2}{*}{\textbf{BART (L)}} 
        & 这款\textbf{衬衫}采用\textbf{清新}的\textbf{蓝色}调，展现出女性内心的纯洁与美好。\textbf{简约}的小\textbf{翻领}设计，修饰出精致小巧的脸型。精致的\textbf{单排扣}门襟，方便穿脱。\textbf{宽松}的\textbf{中长款}版型，对身材的包容性很好。 \\
        & This \textbf{shirt} uses a \textbf{fresh} \textbf{blue} tone to show the purity and beauty of women's hearts. The \textbf{simple} design of small \textbf{lapel} makes the face delicate and small. The exquisite \textbf{single-breasted} placket makes it easy to put on and take off. The \textbf{loose} \textbf{mid-length} clothing design is easily fit the body. \\
      \midrule
      \multirow{2}{*}{\textbf{CPT$_g$ (L)}} 
        & 这款\textbf{衬衫}采用\textbf{中长款}的版型设计，搭配\textbf{宽松}的直筒版型，给身体更多的活动量，行动更为方便，穿着更为舒适。加上\textbf{清新}的\textbf{蓝色}调，更为衬托肤色白皙。结合\textbf{简约}的\textbf{单排扣}门襟，方便穿脱的同时，也丰富了整体的层次感。而精致的小\textbf{翻领}，则为整体增添了几分干练的气质。 \\
        & This \textbf{shirt} adopts a \textbf{mid-length} clothing design with a \textbf{loose} straight type, giving the body more activity, making it more convenient to move and to wear. Coupled with a \textbf{fresh} \textbf{blue} tone, it whitens the skin. Combined with a \textbf{simple} \textbf{single-breasted} placket, it is convenient to put on and take off, which also enriches the overall sense of layers. The exquisite and small \textbf{lapel} adds a capable and experienced style to the whole. \\
      \hline\hline
      \multirow{2}{*}{\textbf{Input}} 
        & [类型, 上衣], [版型, 显瘦], [颜色, 黑色], [衣样式, 毛衣], [衣领型, v领] \\
        & [Type, Top], [Fit, Skinny], [Color, Black], [Clothing Style, Sweater], [Collar Type, V-Neck] \\
      \midrule
      \multirow{2}{*}{\textbf{Reference}} 
        & 金丝的加入为低调内敛的\textbf{黑色}添加闪闪亮色，胸前的\textbf{v领}加排扣设计，摈弃\textbf{毛衣}臃肿松散的固有印象，轻薄\textbf{修身}，保留温度的同时又不至于在猝不及防的倒春寒中失去风度。 \\
        & The golden silk adds sparkling color to the low-key and restrained \textbf{black}. The \textbf{v-neck} on the chest and the breasted design abandons the inherent impression of bloated and loose \textbf{sweaters}, which is light and \textbf{slim}. It keeps the temperature without losing its grace in the unexpected cold spring. \\
      \midrule
      \multirow{2}{*}{\textbf{BART (L)}} 
        & 这款\textbf{黑色}的\textbf{修身}长款\textbf{毛衣}，最大的设计亮点在于衣身\textbf{v领}设计，这样的款式设计使得整件毛衣看起来与众不同，既个性又修身舒适哦。 \\
        & The biggest highlighted design of this \textbf{black} \textbf{slim} long \textbf{sweater} is the \textbf{v-neck} design of the clothing. This design makes the whole sweater unique, personal and comfortable. \\
      \midrule
      \multirow{2}{*}{\textbf{CPT$_g$ (L)}} 
        & 这款\textbf{毛衣}采用了经典的\textbf{黑色}作为底色设计，搭配上\textbf{修身}的版型剪裁，穿起来更加的\textbf{显瘦}。领口处采用\textbf{v领}的剪裁方式，可以起到修饰脸型的作用，更显脸小精致。衣摆处的开叉处理，更是增添了几分随性的感觉。 \\
        & This \textbf{sweater} uses a classic \textbf{black} background with a \textbf{slim} fit cut, which makes you look thin. The neckline adopts a \textbf{v-neck} tailoring method, which can frame the face and make the face small and delicate. The split treatment at the hem adds a casual feel. \\
      \hline\hline
    \end{tabular}
  \end{table}
\end{CJK*}

In addition, the shallow decoder of CPT$_g$ may affect the performance on long text generation. However, the performance gaps are still small. And we believe the multi-task pre-training of CPT closes the gaps. Table~\ref{tab:case-stady-summary} and Table~\ref{tab:case-stady-text} illustrates some examples generated by BART (L) and CPT$_g$ (L). With the help of pre-training for understanding, CPT$_g$ is able to summarize text with more information captured in the input content. 

Moreover, because of the shallow decoder, CPT could generate texts more efficiently (Figure~\ref{fig:speed}), which could be faster than other depth symmetric encoder-decoder Transformers with the same number of layers of the encoder and the decoder.
As BART and CPT have similar number of parameters in both base and large versions. On all generation dataset, the decoding speed of CPT surpass BART with a large margin. Our model achieves $1.4\times \sim 1.5\times$ speedup compared with BART and still maintain comparable generation results in base size. And CPT (L) has up to $1.7\times$ relative speedup compared to BART (L). As a conclusion, the shallow G-Dec is able to speed up the generation with minor performance loss.

\section{Conclusion}

In this paper, we propose CPT, a novel Chinese PTM for both language understanding and generation. With the flexible design, CPT can be assembled and disassembled in various fashions, which could fully exploit the potential of CPT.
Experimental results on a wide range of Chinese NLU and NLG tasks show the effectiveness of CPT.

In future work, we will introduce more specific designs according to Chinese properties, such as better tokenization, pre-training tasks and model architectures.

\Acknowledgements{This work was supported by the National Key Research and Development Program of China (No. 2020AAA0108702), National Natural Science Foundation of China (No. 62022027) and Major Scientific Research Project of Zhejiang Lab (No. 2019KD0AD01).}






\end{document}


\ArticleType{Supplementary File}

\title{Title}{Title for citation}

\author[1]{Aaa AUTHOR}{}
\author[1,2]{Bbb AUTHOR}{{bauthor@xxx.com}}
\author[2]{Ccc AUTHOR}{}
\author[3]{Ddd AUTHOR}{}

\AuthorMark{Author A}

\AuthorCitation{Author A, Author B, Author C, et al}


\address[1]{Affiliation, City {\rm 000000}, Country}
\address[2]{Affiliation, City {\rm 000000}, Country}
\address[3]{Affiliation, City {\rm 000000}, Country}

\maketitle


\begin{appendix}

\section{Importance}
Please use this sample as a guide for preparing your letter. Please read all of the following manuscript preparation instructions carefully and in their entirety. The manuscript must be in good scientific American English; this is the author's responsibility. All files will be submitted through our online electronic submission system at \url{https://mc03.manuscriptcentral.com/scis}.

\section{More information}
The examples at the bottom of the .tex file can help you when preparing your manuscript. We are appreciate your effort to follow our style~\cite{1,2}.

\end{appendix}
